\documentclass{article}

\usepackage{neurips_2023}
\usepackage[x11names]{xcolor}

\usepackage[utf8]{inputenc} % allow utf-8 input
\usepackage[T1]{fontenc}    % use 8-bit T1 fonts
\usepackage{hyperref}       % hyperlinks
\usepackage{graphicx}
\usepackage{enumitem}
\usepackage{amsmath,amssymb}
\usepackage{url}            % simple URL typesetting
\usepackage{booktabs}       % professional-quality tables
\usepackage{amsfonts}       % blackboard math symbols
\usepackage{nicefrac}       % compact symbols for 1/2, etc.
\usepackage{microtype}      % microtypography
\usepackage{xcolor}         % colors
\usepackage{float}
\usepackage{multicol}
\usepackage{multirow}

\title{Enhancing Fast Feed Forward Networks with \\ Load Balancing and a Master Leaf Node}

\usepackage{colortbl}

\author{
  Andreas Charalampopoulos$^1$ \\
  \tt\small{andcharalamp@gmail.com} \\
   \And
   Nikolas Chatzis$^1$\\
   \tt\small{chatznikolas@gmail.com} \\
   \And
   Foivos Ntoulas-Panagiotopoulos$^1$\\
   \tt\small{foivosdoulas@hotmail.gr} \\
   \And
   Charilaos Papaioannou$^1$\\
   \tt\small{cpapaioan@mail.ntua.gr} \\
   \And
   Alexandros Potamianos$^1$\\
   \tt\small{potam@central.ntua.gr} \\
   \And
   $^1$\textnormal{School of ECE, National Technical University of Athens, Greece}
  }
\begin{document}
  % \footnotetext[1]{School of ECE, National Technical University of Athens, Greece}

\maketitle

\begin{abstract}

Fast feedforward networks (FFFs) are a class of neural networks that exploit the observation that different regions of the input space activate distinct subsets of neurons in wide networks. FFFs partition the input space into separate sections using a differentiable binary tree of neurons and during inference descend the binary tree in order to improve computational efficiency.
Inspired by Mixture of Experts (MoE) research, we propose the incorporation of load balancing and Master Leaf techniques into the FFF architecture to improve performance and simplify the training process. We reproduce experiments found in literature and present results on FFF models enhanced using these techniques. The proposed architecture and training recipe achieves up to 16.3\% and 3\% absolute classification accuracy increase in training and test accuracy, respectively, compared to the original FFF architecture. Additionally, we observe a smaller variance in the results compared to those reported in prior research. These findings demonstrate the potential of integrating MoE-inspired techniques into FFFs for developing more accurate and efficient models.

\end{abstract}

\section{Introduction}

Recently, models with billions of parameters have had great success in generative artificial intelligence applications \cite{billion1,billion2,billion3}.  But alongside those impressive results, came the burdensome computational complexity of the \textbf{FeedForward (FF)} layer inference, which is especially present in Transformers\cite{vaswani2023attention}. It has been observed that in wide FF layers, different parts of the input domain activate distinct sets of neurons; this observation can be leveraged to design more efficient models\cite{bengio2016conditional}. As a result the idea of achieving better computational efficiency from sparsely-activated models has gained much attention\cite{gray, gale2020sparse}.

\textbf{Mixture of Experts (MoE)} is an early attempt to take advantage of this sparsity, and continues to be a topic of interest \cite{lepikhin2020gshard,shazeer2017outrageously,MoE}. Recent work on sparsely-activated architectures includes  \textbf{Fast Feed Forward networks (FFF)}\cite{belcak2023fast}. The authors in \cite{belcak2023exponentially,belcak2023fast} indicate that FFFs can be used instead of vanilla FF and MoE architectures in transformers and \textbf{Large Language Models (LLM)} without incurring any significant loss in accuracy, while realizing a considerable speed-up during inference. Inference acceleration in FFFs is achieved through a tree-conditional activation of neurons.

While trying to reproduce experiments from \cite{belcak2023fast}, we verified that FFFs suffer from training instability. This can be also inferred from the large variance in results that are reported also in Table 5 of \cite{belcak2023fast}, where the variance among identical training runs is high. Further we observed that certain subtrees in the FFF architecture were activated significantly more than others during inference, i.e., there was significant imbalance on the utilization of the FFF. To address these two issues and motivated by the MoE literature \cite{dai2024deepseekmoe}, we propose two modifications to the FFF architecture: 1) introducing load balancing to better utilize all FFF subtrees, and 2) adding a master leaf node in parallel to the FFF topology that contributes to the output with a constant mixture coefficient, so that input sequences that cause ``wider'' neural activation patterns can be better serviced. We show that the proposed enhancements improve classification performance on the MNIST and FashionMNIST datasets. Further we show that the enhanced FFFs achieve better overall training stability compared to vanilla FFFs.

Our contributions can be summarized as follows:
\begin{enumerate}
    \item We propose an enhanced FFF architecture (eFFF) that incorporates a load balancing term at the loss function and a master leaf node that gets linearly mixed with the FFF output. 
    \item We provide experimental validation on the MNIST and FashionMNIST datasets showing that the proposed method yields better classification accuracy both during training and testing, and leads to more stable training runs (reduced variance). Further, we perform ablation experiments showing the contribution of each proposed enhancement.  
    \item We also provide all the code necessary to reproduce our experiments in the following GitHub repository\footnote[2]{\url{https://github.com/AndreasCharalamp/fastfeedforward-experiments}}.
\end{enumerate}

\section{Related Work}
The importance of inference speedup in feedforward neural networks is widely recognised and several approaches have been proposed. Recent works have successfully managed to reduce the feedforward layer inference time. The Mixture of Experts (MoE) approach,  as explored in Shazeer et al. (2017) \cite{shazeer2017outrageously}, has demonstrated its effectiveness towards inference speedup. MoE involves dividing the feedforward layer into distinct sets of neurons known as ``experts'', with a gating layer trained to select which mixture of experts to utilize during the forward pass. This method enhances inference speed by utilizing only the top-performing $k$ blocks, or a similar variation thereof. It effectively reduces inference time by a constant factor while maintaining a linear relationship with the width of the feedforward layer. However, it depends on noisy gating to balance the load among the experts, adding complexity to the training process and encouraging redundancy.

In \cite{belcak2023fast}, %Peter Belcak and Roger Wattenhofer 
the authors introduced the Fast Feedforward (FFF) architecture as an alternative to the feedforward (FF) architecture. FFF operates by accessing blocks of its neurons in logarithmic time, offering improved efficiency. It accomplishes this by dividing the input space into separate regions using a differentiable binary tree, simultaneously learning the boundaries of these regions and the neural blocks assigned to them. Neurons are executed conditionally based on the tree structure during inference: a subset of node neurons determines the mixtures of leaf neuron blocks required to generate the final output.
Further in \cite{belcak2023fast,belcak2023exponentially}, the authors demonstrate that FFFs can be up to 220 times faster than feedforward networks and up to 6 times faster than mixture-of-experts networks. Additionally, the authors claim that FFFs exhibit superior training properties compared to mixture-of-experts networks due to their noiseless conditional execution approach.

In this paper, we utilize the concept of load balancing, previously introduced in MoE\cite{MoE,shazeer2018meshtensorflow,lepikhin2020gshard}, to ensure a balanced load across FFF's leaves, aiming to improve training stability. In the context of MoE,\cite{shazeer2017outrageously} an additional term in the loss function is introduced, in order to encourage experts to receive roughly equal numbers of training examples. This idea proves to be significant for load balancing purposes on distributed hardware.

Furthermore, we propose mixing the FFF's output with that of another neural network with much fewer neurons. We call this network ``master leaf'' as it is similar to the leaves of FFF. The weight of the output of the master leaf is set to be a trainable parameter. Inspiration for this was drawn from\cite{master2}, where authors proposed enhancing MoE performance by integrating a base network alongside the selected expert. This is shown to not only improves model accuracy, but also provides an early exit output during inference, reducing computational redundancy for ``easier'' samples. Additionally, computational efficiency is achieved by reusing early layers of the base model as inputs to the gate and the experts.

\section{Method}
\subsection{FFF architecture}
Fast feedforward networks (FFFs) are designed to capitalize on the phenomenon wherein different parts of the input domain activate distinct sets of neurons in wide networks. FFFs partition the input space into separate sections using a differentiable binary tree, enabling the concurrent learning of both the boundaries delineating these sections and the neural units associated with them. This is accomplished through the tree-conditional activation of neurons: a designated subset of node neurons determines the combinations of leaf neuron blocks to be computed for generating the output.

\subsection{Training Process}
The nodes are arranged in a differentiable tree that makes a soft choice over the leaves in the 
form of a stochastic vector. 
In training, FFF performs a mixture of experts over all leaves in $\mathcal{L}$, where $\mathcal{L}$ is the set of leaves, with the weights of the 
mixture computed by ascending through the tree from the root node. During inference, the 
decision at each node is taken to be the greatest weight, and the forward pass algorithm 
proceeds from the root, always choosing only {\bf one branch} depending on the local node 
decision. All leaves are simple Feed-Forward (FF) networks with one hidden layer of width $\ell$ , and ReLU (Rectified Linear Unit) activation function. The nodes of the tree are simple neurons that use sigmoid activation function.
Following the notation of \cite{belcak2023fast} we will refer to the total number of neurons in each model (excluding the tree nodes in an FFF) as the \textbf{training width} and will denote it as $w$. The number of neurons of each leaf will be denoted by $\ell$ and we will call it \textbf{leaf width}. 
The output of an FFF during training is of the following form:
\begin{equation}
\begin{split}
    F\!F\!F_{\text{train}}(x) = \sum_{1\le i \le |\mathcal{L}|}l_{i}(x)\, c_{i}(x),\end{split}
\end{equation}
where $\sum_{1\le i \le |\mathcal{L}|}c_{i}(x) = 1$,  $|\mathcal{L}|$ is the number of  leaves, $\ell_{i}(x)$ is the output of leaf $i$ and $c_i(x)$ is the mixture coefficient of leaf $i$ computed as the product of the edges in the path from the root to each leaf $l_{i}$ as shown in Fig.~1.
    \begin{figure}[H]
    \centering
    {\includegraphics[width=1\textwidth]{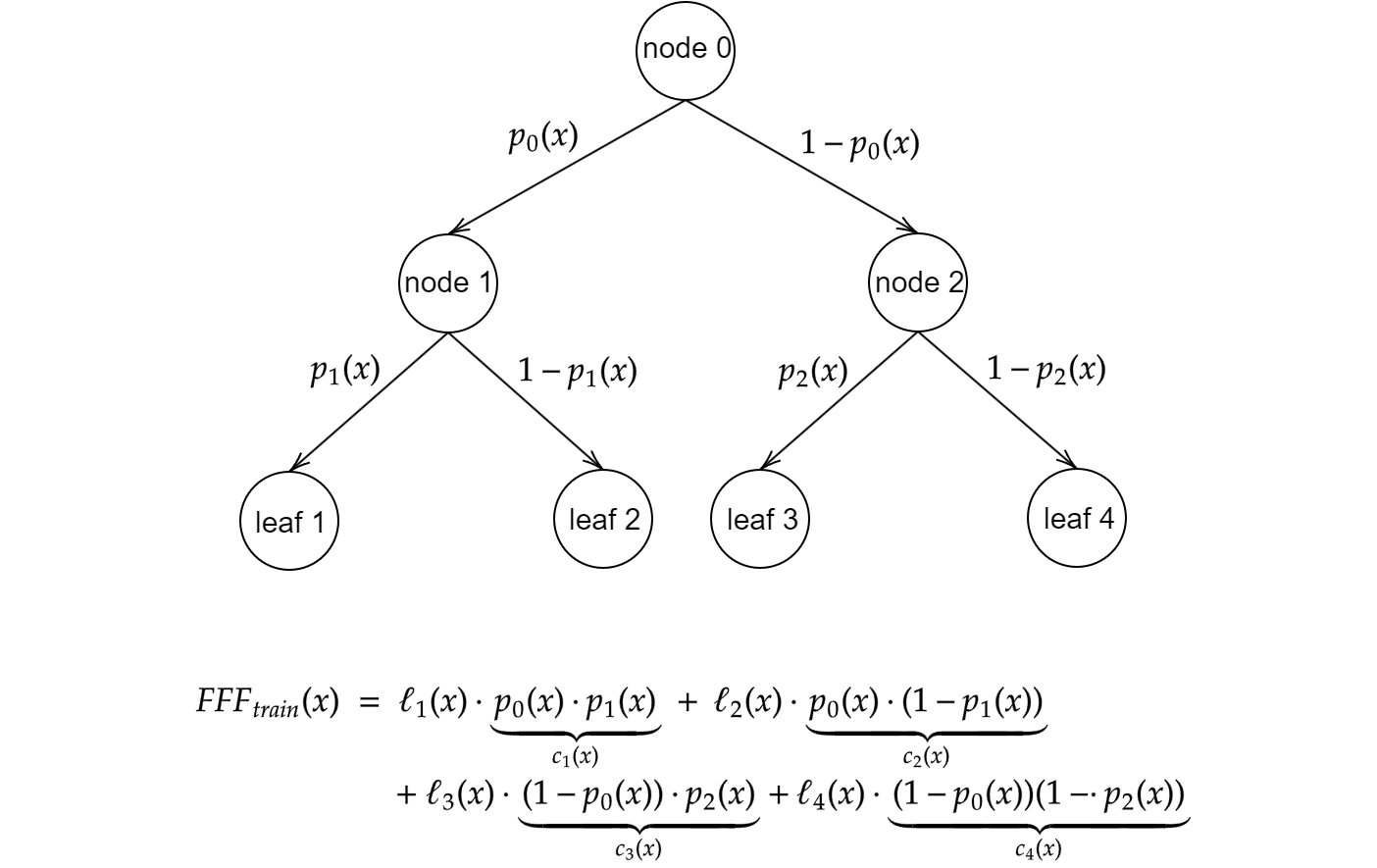} }%
    \vspace*{-3mm}
    \caption{Visualization of FFF training  for tree depth 2.}
    \label{fig:fore-back}%
\end{figure}
During inference the output is computed by taking hard decisions at each level of the hierarchy resulting in only $c_*$ of the $c_i$ being 1 and the rest being 0, i.e.,
\begin{equation}
    \begin{split}
        F\!F\!F_{\text{inference}}(x) = l^{*}(x),
    \end{split}
\end{equation}
where $l^{*}$ is the leaf that we end up on, following the edges of greater value. This way, even though $2^{d}\cdot \ell + 2^{d}-1$ neurons are used for training, where $d$ is the depth of the tree, only $\ell + d-1$ are used for inference.

In \cite{belcak2023fast} the following loss function is used:
\begin{equation*}
    \begin{split}
        L = L_{\text{pred}} + h\,  L_{\text{harden}},
    \end{split}
\end{equation*}
where $L_{\text{pred}}$ is the task cross entropy loss, $L_{\text{harden}}$ is a term that pushed the decisions at each level of the tree to be either 0 or 1 and $h$ is the training hyperparameter controlling the effect of the hardening. Specifically, $L_{\text{harden}}$ is defined as:
\begin{equation*}
    \begin{split}
        L_{\text{harden}} = \sum_{i\in \mathcal{B}}\sum_{N\in \mathcal{N}}H(N(i)),
    \end{split}
\end{equation*}
where $\mathcal{B} $ is a batch of samples, $\mathcal{N}$ is the set of tree nodes of the FFF, $H(p)$ the entropy of a
Bernoulli random variable $p$. This extra term is needed so that all edges of the tree have values close to $1$ or $0$ for all inputs. The hardening term is important because the
FFF is trained to output predictions in the form of a weighted sum of its leaves, while during inference we make hard 0 vs 1 decision while descending the tree. In order for inference output $F\!F\!F_{\text{inference}}(x)$ 
 to be as close as possible to training ouput $F\!F\!F_{\text{train}}(x)$ (see Eqs.~(1) and~(2) above) we  aim for all $c_{i}$ to be near $0$ and only $c_*$ to be close to 1.

 Thus, through the hardening term, we seek to force the weight of leaf $l^{*}$ to be close to $1$ and the weights of the rest of the leaves to be close to 0.
\subsection{Load Balancing}
During our training trials with FFFs we noted that they are highly sensitive to poor initialization of weights. This is evident from the significant variability in test accuracy observed across multiple runs of the same training procedure. Similar challenges are also noted in \cite{belcak2023fast}, particularly in the Table~4 in the Appendix, where accuracy variations are documented. To elaborate further, the loss function does not promote a wide usage of the leaves. Consequently, during training, if a leaf is assigned to a region of little relevance, it is likely to complete the training process without effectively capturing any meaningful representation.

To tackle this, we study how this problem was addressed in MoE architectures. Following the idea from \cite{MoE} we propose to add the following term into the loss function:
       \begin{equation*}
       \begin{split}
          L_{\text{balance}} =   2^{d}\, \sum_{i \in \text{leaves}}f_i \, P_i,
       \end{split}
   \end{equation*}
where $f_i$ is the fraction of the inputs dispatched to leaf $l_i$ and $P_i = \frac{1}{|\mathcal{B}|}\sum_{x\in \mathcal{B}}c_{i}(x)$ is the sum of the coefficients of each leaf $i$ on the current batch $\mathcal{B}$.The term $L_{\text{balance}}$ is minimized when the load is evenly balanced on all leaves. The resulting total loss $L'$ is now
\begin{equation*}
    \begin{split}
        L' = L_{\text{pred}} + h\,  L_{\text{harden}} + \alpha \, L_{\text{balance}},
    \end{split}
\end{equation*}
where $\alpha$ is a hyperparameter  controlling the effect of the load balancing term.

\subsection{Master Leaf}
Inspired from \cite{master2}, we experiment with the addition of an extra neural component. Instead of allowing each partition set of the input space to be processed exclusively by independent sets of neurons (leaves) during inference, we provide an additional set of neurons which contributes to the output for all inputs, and not only a subset of them like the rest of the leaves.
We introduce a master leaf, that contributes to the final output with a factor $k$. During {training}, the output of the new architecture is formulated as follows:
\begin{equation*}
    \begin{split}
      F\!F\!F_{\text{ML}_\text{Train}}(x)  
      = k\sum_{1\le i \le |\mathcal{L}|}l_{i}(x)\, c_{i}(x) + (1-k)\, M\!L(x),
    \end{split}
\end{equation*}
where $|\mathcal{L}|$ is the number of the leaves, $\ell_{i}(x)$ is the output of leaf $i$, $c_i(x)$ is the mixture coefficient of leaf $i$, $M\!L$ is the output of the master leaf and $k$ is a trainable parameter with $0<k<1$. This linear fusion method is further elucidated in Fig.~2.
\begin{figure}[H]
    \centering    {\includegraphics[width=1\textwidth]{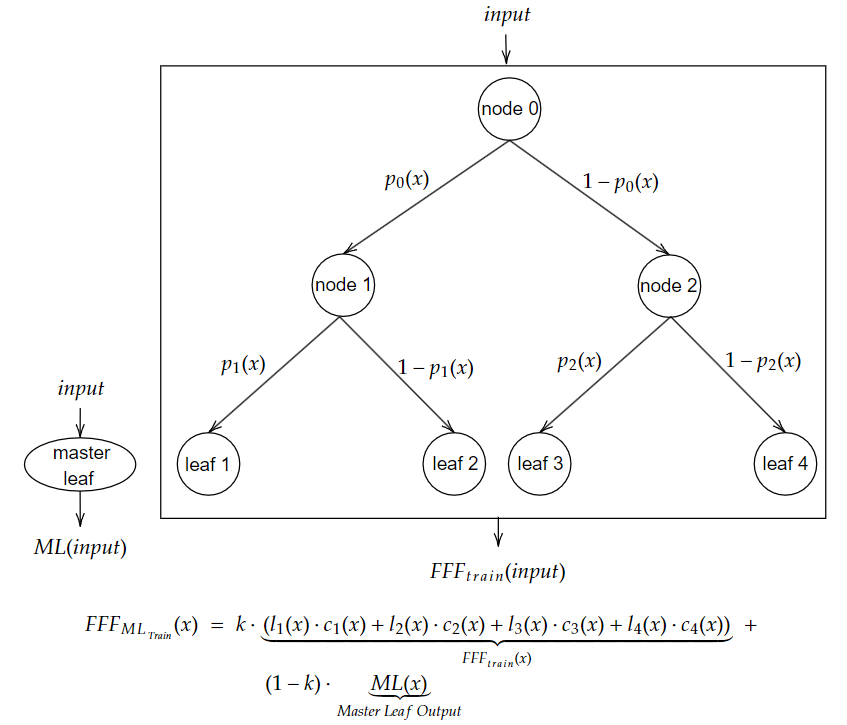} }%
        \vspace*{-3mm}
    \caption{Visualization of FFF training with master leaf architecture.}
    \label{fig:masterleaf}%
\end{figure}
During \textbf{inference}, the output of the new architecture is formulated as follows:
\begin{equation*}
    \begin{split}        
    FFF_{\text{ML}_{\text{Inference}}}(x) 
    = 
    k\,\ell^{*}(x) + (1-k)\, M\!L(x),
    \end{split}
\end{equation*}
where $ \ell^{*}(x) $ is the output of the leaf with the greatest mixture coefficient $c^{*}(x)$.

The master leaf undergoes training concurrently with the FFF on the entire dataset. Each FFF leaf is tasked with handling a distinct subset of the input space. Consequently, the introduction of the master leaf enriches the ``localized'' output of a leaf through the incorporation of the well-trained feedforward network output\footnote{The  master leaf output can be calculated in parallel with the output of the leaf chosen from the FFF. Consequently, with proper implementation, it should not significantly affect inference speed.}.

\section{Experimental Setup}
We conduct a series of experiments to investigate the benefits in performance resulting from:
\begin{enumerate}[label = (\arabic*)]
    \item the inclusion of the load balancing term in the loss function and
    \item the integration of the output of an FFF with the master leaf output, as described above.
\end{enumerate}

Building upon the foundation laid in \cite{belcak2023fast}, we adopt training and test accuracy as our evaluation metrics to facilitate direct comparison with the literature. Each experiment focuses on image classification, with classification accuracy assessed through the softmax of output logits in the usual way. Results are reported on the MNIST and FashionMNIST image classification databases. The reader can refer to \cite{belcak2023fast} for details on the database and experimental setup, which are mirrored here. 

\subsection{Experiments 1 and 2: Load Balancing}

In order to investigate the effect of load balancing we reproduce the experiment from Table 1 in \cite{belcak2023fast} (referred henceforth as {\bf baseline}) and compare the performance when using the load balancing term in the loss function (referred henceforth as {\bf balanced}). We report classification accuracy on the MNIST and FashionMNIST datasets for the following sets of parameters in experiment 1: leaf width $l\in\{8,4,2,1\}$ and training width $w=16$. We train for $300$ epochs with learning rate $lr = 0.001$, loss hyperparameters $h=1, \alpha =1$  and another $300$ epochs with $lr= 0.001, h=3, \alpha = 0$. We use the Adam optimizer and early stopping (if no increase in loss is observed over 50 epochs).

Additionally in experiment 2,  we explore cases for the FashionMNIST database where training width is $w=128$, $l\in\{8,4,2,1\}$ and also $l\in\{64,32,16\}$ that were not included in the initial study. This allows us to observe the accuracy attained when the leaf size approaches that of a simple feedforward network.

We perform 10 training runs and report  best accuracy and worst accuracy in Tables \ref{tab:tab1} and \ref{tab:tab2}.

\subsection{ Experiment 3: Master Leaf with Load Balancing}

Next, we investigate the performance of Master Leaf architecture on the MNIST dataset. For this experiment we fix the master leaf size at $8$ and also include the load balancing term in the loss function (henceforth referred to as ``{\bf master leaf + balanced}''). Training takes place for $200$ epochs with $lr = 0.001, h=1, \alpha =1$  and another $100$ epochs with $lr= 0.001, h=3, \alpha = 0$. We train using the Adam optimizer and early stopping (if no increase is observed for over $50$ epochs). 
We perform 5 training runs and report  best accuracy and worst accuracy in
Table \ref{tab:tab3}.

We publish the parameters for all  trained models in our GitHub repository (see link in Introduction).

\section{Experimental Results}
\subsection{Experiment 1: Load Balancing}
\begin{table}[htbp]
\centering
\begin{tabular}{|c|c|c|c|c|c|}
\hline    \multicolumn{5}{|c|}{ MNIST } \\
\hline  \multicolumn{5}{|c|}{$w=16$}  \\
\hline &  \multicolumn{2}{c|}{train accuracy} & \multicolumn{2}{c|}{test accuracy} \\ \hline
 &  baseline & balanced & baseline & balanced \\
\hline vanilla FF  & \cellcolor{Snow2}{$98.0 \pm 0.9$} & - & \cellcolor{Snow2} $95.2 \pm 0.5$ & -   \\
 $\ell=8$ & \cellcolor{Snow2}$94.6 \pm 19.5 $&  $94.6 \pm 7.0 $&\cellcolor{Snow2} $93.1 \pm 16.6$ & $93.5 \pm 6.1$  \\
 $\ell=4$ & \cellcolor{Snow2}$91.6  \pm 29.3$ & $94.2 \pm 3.9$ &\cellcolor{Snow2} $90.8 \pm 27.2$ & $91.3 \pm 8.9$  \\
$\ell=2$ &\cellcolor{Snow2} $92.1 \pm 7.3$ & $95.0 \pm 1.7$ &\cellcolor{Snow2} $90.3 \pm 5.6$  & $91.0 \pm 2.7$\\
  $\ell=1$ &\cellcolor{Snow2} $91.7 \pm 7.4$ & $95.2 \pm 3.0$ &\cellcolor{Snow2} $89.9 \pm 6.4$ & $89.0 \pm 8.1$   \\
\hline \hline    \multicolumn{5}{|c|}{ FashionMNIST } \\
\hline  \multicolumn{5}{|c|}{$w=16$} \\
\hline &  \multicolumn{2}{c|}{train accuracy} & \multicolumn{2}{c|}{test accuracy}  \\
\hline &  baseline & balanced & baseline & balanced \\
\hline vanilla FF  & \cellcolor{Snow2}$91.0 \pm 0.7$ & - &\cellcolor{Snow2}  $86.4 \pm 0.4$ &  -   \\
  $\ell=8$ & \cellcolor{Snow2}$86.7 \pm 12.1$ & $90\pm1.5$ &\cellcolor{Snow2} $84.2 \pm 10.9$ &  $86.1 \pm 1.1$  \\
 $\ell=4$ & \cellcolor{Snow2}$86.4 \pm 25.0$ & $89.5 \pm 0.6$ & \cellcolor{Snow2}$83.3 \pm 24.5$ &  $85.8\pm 0.9$ \\
  $\ell=2$ &\cellcolor{Snow2} $84.5 \pm 21.0$ &  $91.2\pm 1.4$ & \cellcolor{Snow2}$83.0 \pm 11.0$ & $85.4 \pm 2.5$   \\
  $\ell=1$ & \cellcolor{Snow2}$79.7 \pm 9.0$ &$92.7 \pm 1.6$ & \cellcolor{Snow2}$78.4 \pm 8.0$ & $80.3 \pm 9.1$  \\
\hline
\end{tabular}
\vspace*{2mm}
\caption{Training and test image classification accuracy of baseline and  load balanced models on MNIST and FashionMNIST. $w$ is the training width, $\ell$ is the leaf width. Results with grey background are copied from \cite{belcak2023fast} for comparison.  $x\pm y$ means that, from the 10 training runs best accuracy was $x$ and worst was $x-y$.}
\label{tab:tab1}%
\end{table}

The results for the baseline FFF model as reported in \cite{belcak2023fast} and the load balanced FFF model are shown in Table~1 for the MNIST and FashionMNIST datasets. 
The load balanced FFF model with the proposed training strategy outperforms the baseline in all settings. Specifically, we observe an increase in training accuracy up to $16.3\%$ absolute, achieved for $\ell = 1$ for FashionMNIST, while the test accuracy exhibits a maximum increase of $3.0\%$, achieved for $\ell = 4$ for FashionMNIST. The average absolute training accuracy improvement for MNIST is  $2.3\%$ that translates to $27\%$ relative error reduction. Test accuracy improvement is small typically  $0.5\%$ absolute for MNIST, but consistent and significant for FashionMNIST on average $2.2\%$ absolute and $10\%$ relative error rate reduction. 

Moreover, it is apparent that accuracy  variability among training runs has diminished by 4 to 5 times on average for both training and testing when using load balancing. However, accuracy variability  remains significantly higher than for vanilla FFs. We believe variance in deep models remained high because asking our model to partition MNIST and FashionMNIST into $w=16$ meaningful regions might lead to overfragmentation of the input space, as explained in \cite{belcak2023fast}. One last thing to note is that the load balancing term appears to introduce overfitting  especially for deeper models, i.e., the training accuracy improves faster than the test accuracy. 

\subsection{ Experiment 2: Load Balancing with Larger Training and Leaf Width}
The increase in accuracy is made more apparent via Table~2 where we present results also for $w=128$ case for FashionMNIST and also for deeper models.

\begin{table}[htbp]
\centering
\resizebox{\textwidth}{!}{
\begin{tabular}{|c|c|c|c|c|c|c|c|c|c|}
\hline     \multicolumn{9}{|c|}{ FashionMNIST } \\
\hline & \multicolumn{4}{c|}{$w=16$} &  \multicolumn{4}{c|}{$w=128$} \\
\hline &  \multicolumn{2}{c|}{train accuracy} & \multicolumn{2}{c|}{test accuracy} & \multicolumn{2}{c|}{train accuracy} & \multicolumn{2}{c|}{test accuracy} \\
\hline &  baseline & balanced & baseline & balanced &baseline & balanced & baseline & balanced \\
\hline vanilla FF  & \cellcolor{Snow2}91.0 & - &\cellcolor{Snow2}  86.4 & -  & \cellcolor{Snow2}99.3 & -  &\cellcolor{Snow2} 89.6 & -  \\
   $\ell=64$ &- & -& - &-  & 95.6 & 97.0 & 88.8 &88.9 \\
   $\ell=32$ & -& -& -&-  & 93.1 & 96.5 &87.9 &88.2 \\
   $\ell=16$ & - & - & - & -  & 92.5 & 94.3 & 87.1 & 87.5 \\
   $\ell=8$ & \cellcolor{Snow2}86.7 &  90.0 &\cellcolor{Snow2} 84.2 & 86.1  &\cellcolor{Snow2} 90.5 & 92.8 &\cellcolor{Snow2} 86.1 &86.7 \\
  $\ell=4$ & \cellcolor{Snow2}86.4 &89.5 & \cellcolor{Snow2}83.3 &  85.8  &\cellcolor{Snow2} 89.0 & 89.6 &\cellcolor{Snow2} 85.4 & 85.8 \\
  $\ell=2$ &\cellcolor{Snow2} 84.5 & 91.2 & \cellcolor{Snow2}83.0 & 85.4  &\cellcolor{Snow2} 87.3 & 88.3 &\cellcolor{Snow2} 84.3 & 84.8  \\
 $\ell=1$ & \cellcolor{Snow2}79.7 & 92.7 & \cellcolor{Snow2}78.4 & 80.3  &\cellcolor{Snow2} 78.7 & 84.5 &\cellcolor{Snow2} 77.7 & 79.9  \\
\hline
\end{tabular}
}
\vspace*{2mm}
\caption{Training and test accuracy attained with load balancing (baseline vs balanced) for the FashionMNIST database and for larger training  and leaf widths. Baseline results that are copied from \cite{belcak2023fast} are highlighted in grey, baseline results for $\ell=16, 32, 64$ are our own.}
\label{tab:tab2}%
\end{table}
 We observe that load balancing improves the accuracy over the baseline FFF model for all setups. For $\ell \in \{1,2\}$ we observe that we have better results for the $w=16$ case rather than $w=128$ probably due to input space overfragmentation mentioned before. Results could be further improved if we harden our models for more epochs. Note that load balancing provides consistent accuracy improvement even for best performing deep models. The more the leaves in the model, the harder it is to find a good partition of the input space without using load balancing.
 
\subsection{Experiment 3: Master Leaf and Load Balancing}

Results on MNIST when adding a master leaf node of size $8$ are shown in Table~3. Compared to the baseline and Table~1 performance (using load balancing only) we see significant improvement on training accuracy both for $w=16$ and $w=128$. Test accuracy also improves in the vast majority of the cases. As expected, the improvement is greater for $w=16$ than for $w=128$, typically $3.8\%$ vs $1.3\%$ absolute accuracy improvement, respectively. Additionally, adding the master leaf   further reduces the performance variability among runs bringing it to reasonable levels comparable to vanilla FF for  $w=16$. Overall, mixing the output of an FFF with the output of a simple neural network is a very promising direction.
\newpage
\begin{table}[htb]
\centering
\begin{tabular}{|c|c|c|c|c|c|}
\hline     \multicolumn{5}{|c|}{ MNIST } \\
\hline  \multicolumn{5}{|c|}{$w=16$} \\
\hline &  \multicolumn{2}{c|}{train accuracy} & \multicolumn{2}{c|}{test accuracy} \\
\hline & baseline & master leaf + balanced & baseline & master leaf + balanced \\
\hline vanilla FF  & \cellcolor{Snow2}$98.0 \pm 0.9$ & - &\cellcolor{Snow2}  $95.2 \pm 0.5$ & -   \\
   $\ell=8$ & \cellcolor{Snow2}$94.6 \pm 19.5$ &  $96.7 \pm 1.4$ &\cellcolor{Snow2} $93.1 \pm 16.6$ & $94.8 \pm 0.5$   \\
  $\ell=4$ & \cellcolor{Snow2}$91.6 \pm 29.3$ & $96.7 \pm 1.6$ & \cellcolor{Snow2}$90.8 \pm 27.2$ &  $94.7 \pm 2.0 $  \\
  $\ell=2$ &\cellcolor{Snow2} $92.1 \pm 7.3$ & $97.2 \pm 1.5$ & \cellcolor{Snow2}$90.3 \pm 5.6 $ & $94.1 \pm 1.1$  \\
  $\ell=1$ & \cellcolor{Snow2}$91.7 \pm 7.4$ & $97.3 \pm 0.9$ & \cellcolor{Snow2}$89.9 \pm 6.4$ & $93.8 \pm 1.8$    \\
\hline \hline   \multicolumn{5}{|c|}{$w=128$} \\
\hline &  \multicolumn{2}{c|}{train accuracy} & \multicolumn{2}{c|}{test accuracy} \\
\hline & baseline & master leaf + balanced & baseline & master leaf + balanced \\
\hline vanilla FF  & \cellcolor{Snow2}$100 \pm 0.0 $ & -  &\cellcolor{Snow2} $98.1 \pm 0.1$ & -  \\
   $\ell=8$ & \cellcolor{Snow2} $99.3 \pm 1.0$ & $100 \pm 0.0$ &\cellcolor{Snow2} $94.9 \pm 0.6$ & $95.1 \pm 0.3$ \\
  $\ell=4$ & \cellcolor{Snow2} $97.6 \pm 0.6$ & $99.8 \pm 0.5$ &\cellcolor{Snow2} $93.6 \pm 0.5$ & $95.0 \pm 1.8$ \\
  $\ell=2$ &\cellcolor{Snow2} $96.2 \pm 1.4$ & $99.7 \pm 2.6$ &\cellcolor{Snow2} $92.4 \pm 0.6$ & $93.7 \pm 3.1$  \\
  $\ell=1$ &\cellcolor{Snow2} $94.1 \pm 0.9$ & $99.7 \pm 0.7$ &\cellcolor{Snow2} $92.0 \pm 0.7$ & $91.6 \pm 10.1$ \\
\hline
\end{tabular}
\vspace*{2mm}
\caption{Training and test accuracy attained with master leaf models also using the load balancing loss term for the MNIST database. $w$ is the training width, $\ell$ is the leaf width. Baseline results copied from \cite{belcak2023fast} are highlighted in grey. $x\pm y$ means that from the 5 training runs best accuracy was $x$ and worst was $x-y$.}
\label{tab:tab3}%
\end{table}
\section{Conclusions}
We enhanced the FFF architecture proposed in \cite{belcak2023fast} with a load balancing loss term and a master leaf node achieving consistently improved accuracy for the MNIST and FashionMNIST image classification tasks. %In more detail, training accuracy increased up to $14.3\%$ and test accuracy increased up to $3.0\%$. 
Particularly noteworthy is the increase in accuracy for deep FFFs. 
%We attribute this phenomenon to the proliferation of leaves, which amplifies the likelihood of encountering local optima where only a subset of leaves is utilized.
Equally noteworthy is the reduction in accuracy variability across our training runs. This result underscores the robustness conferred by the incorporation of the load balancing term and master leaf architecture into FFFs.
%Furthermore, we explored the concept of employing a master leaf, which contributes to formulating the output with a constant mixture coefficient for any input.
%We believe that it is possible to implement Master Leaf computations in parallel with the output of FFF. Thus, adding a Master Leaf should affect minimally the speed-up gains of FFFs vs traditional NN architectures.
The main conclusions from the 3 experiments and proposed future directions are discussed next:
\begin{enumerate}
    \item Experiment  1 results confirm our belief that the largely varying test accuracy are caused by unbalanced trees. Adding the load balancing term in our training we achieve better leaf utilization resulting in increased robustness.  
    \item Experiment  2 results indicate that we can achieve significantly better performance using leaf balance, as we surpassed \cite{belcak2023fast} best accuracy for all $w,\ell$. Thus, we believe it is worth evaluating the performance using more training epochs and a larger range of  parameters  to fully explore the potential of the method.
    \item Experiment  3 results show that  the master leaf architecture outperforms FFF models, in terms of test and train accuracy, for all cases  investigated. Expanding these experiments to other datasets and exploring various values of master leaf width holds significant potential for further performance improvements.  
\end{enumerate}

\vspace*{1mm}
{\bf Limitations:} We did not explore fully the (hyper-) parameter space due to computational resource limitations; it is possible that results can be further improved via parameter tuning.

\section*{Acknowledgements}
We wish to thank the authors of \cite{belcak2023fast} for their guidance on the FFF implementation.  This work was part of a term project for the  Pattern Recognition class of the ECE curriculum at NTUA. 

\newpage

\bibliographystyle{IEEEtran} % We choose the "plain" reference style
\bibliography{paper2} % Entries are in the refs.bib file
%%%%%%%%%%%%%%%%%%%%%%%%%%%%%%%%%%%%%%%%%%%%%%%%%%%%%%%%%%%%

\end{document}